**Meta-data Study in Autism Spectrum Disorder Classification Based on Structural MRI**

Ruimin Ma, Yanlin Wang, Yanjie Wei*, Yi Pan*

Shenzhen Institute of Advanced Technology, Chinese Academy of Sciences, Center for High Performance Computing, Shenzhen 518055, Guangdong, China

*Corresponding emails: yi.pan@siat.ac.cn, yj.wei@siat.ac.cn

## ABSTRACT

Accurate diagnosis of autism spectrum disorder (ASD) based on neuroimaging data has significant implications, as extracting useful information from neuroimaging data for ASD detection is challenging. Even though machine learning techniques have been leveraged to improve the information extraction from neuroimaging data, the varying data quality caused by different meta-data conditions (i.e., data collection strategies) limits the effective information that can be extracted, thus leading to data-dependent predictive accuracies in ASD detection, which can be worse than random guess in some cases. In this work, we systematically investigate the impact of three kinds of meta-data on the predictive accuracy of classifying ASD based on structural MRI collected from 20 different sites, where meta-data conditions vary. Data size is shown to have a negative impact on the predictive accuracies, indicating that simply adding more data into training won't help, unless the data quality is strictly controlled. The phenotypic data of participants, like age, gender, and eye status during rest of scan, are investigated as well, which show weak correlations with predictive accuracies, suggesting that different phenotype have insignificant impact on the structure of brain regions that are affected with ASD. Finally, the scan parameters for collecting structural MRI are explored, which show vague relationship with predictive accuracies, saying that we can't improve the quality of structural MRI merely based on the scan parameters. Our demonstrated meta-data study may advance the data collection strategies in the neuroimaging field and indirectly improve the accuracy of ASD detection based on neuroimaging data in the future.

## INTRODUCTION

Accurate diagnosis of autism spectrum disorder (ASD) based on neuroimaging data becomes popular and critical nowadays.[1, 2] However, the intricate structure and function of the brain makes it challenging to observe useful information from the neuroimaging data, mainly due to its low-resolution nature, which is usually impacted by the imaging hardware, scanning time, and the cost.[3] Recently, artificial intelligence (AI) makes great achievements in pattern recognition of scientific data, like plasma control,[4] protein design,[5] and materials discovery,[6-8] which is naturally being considered as an assistive approach for studying neuroimaging data. Kong et al. adopted deep learning techniques for classifying ASD based on 182 structural MRI (s-MRI) data collected from ABIDE-I, achieving a predictive accuracy of 0.90 in the binary classification.[9] Gazzar et al. leveraged 1-d convolutional networks for classifying ASD based on 2085 resting-state functional MRI data collected from ABIDE I+II, achieving a predictive accuracy of 0.68 in the binary classification.[10] However, as indicated by the above research, even assisted by the AI approaches, achieving high predictive accuracy in ASD classification based on neuroimaging data can't be secured. According to a recent survey in diagnosing ASD based on neuroimaging data and AI techniques,[11] the predictive accuracy varied significantly as training dataset varied.

As AI research matures, researchers notice that advancing AI algorithms plays less and less critical role in applying AI to real-world scenarios, and shift the focus gradually to data-centric AI, where developing strategies to acquire high-quality data matters most.[12] Acquiring data from different sources brings noise to the data at the meantime, thus the meta-data that describes the data's characteristics are important, based on which the researchers can develop data collection strategies. Google Dataset Search[13] is a good example in collecting metadata for data, where researchers can search for the data based on the metadata in the "long tail" of the web. In the health care field, process mining approach is leveraged to aggregate the metadata for patients, based on which the data quality issues can be well addressed later.[14] When it comes to the neuroimaging field, detailed metadata are also recorded for some benchmark databases, like ABIDE-I[15] and ABIDE-II[16].

Meta-data analysis is important for AI research in detecting ASD based on neuroimaging data, where the collected data are usually heterogeneous.[15, 16] Nielsen et al. quantified the relationship between ASD and functional connectivity MRI and found that the classification accuracy was much higher for the model trained with data where BOLD imaging times were longer.[17] Katuwal et al. quantified the relationship between ASD and s-MRI and found that the classification accuracy can be improved when adding the meta-data, such as age and IQ, into the training.[18] Vigneshwaran quantified the relationship between ASD and s-MRI and found an age-dependent effect in the data, where the classification accuracy was higher in adult group than that in adolescent group.[19] Although the effects of different metadata were explored in classifying ASD based on AI and neuroimaging data, a systematic study of meta-data and their impacts on the accuracy of classifying ASD is still lacking.

In this work, we systematically investigate the metadata and their impacts on the accuracy of classifying ASD based on the s-MRI. 1100 s-MRI from 20 different sites are collected from ABIDE-I, and their corresponding metadata are explicitly recorded as well. The volume-based features are extracted via FreeSurfer from s-MRI for classifying ASD. The genetic algorithm is leveraged for feature selection, which is conducted in a hierarchical way, to eliminate the irrelevant features for classifying ASD. After the feature selection, the predictive accuracies of ASD classification based on selected features are calculated for 20 different sites. The relationship between predictive accuracy and three kinds of metadata, including data size, phenotypic data of participants, and scan parameters for collecting s-MRI, are explored in this work, which gives useful implications on how to acquire high quality s-MRI for ASD detection in the future.

## METHODS

**Dataset.** 1100 s-MRI data from 20 different sites (UM1, UM2, CMU, CalTech, KKI, NYU, MAX_MUN, LEUVEN1, LEUVEN2, OHSU, OLIN, PITT, STANFORD, SDSU, SBL, TRINITY, USM, UCLA1, UCLA2, YALE) are obtained from the ABIDE-I,[15] with 530 of them belonging to ASD participants and 570 of them belonging to neurotypical (NT) participants. Detailed description of participant recruitment protocol and s-MRI data collection can be checked at http://fcon_1000.projects.nitrc.org/indi/abide. The s-MRI data are preprocessed using FreeSurfer software (version 6.0.0, http://surfer.nmr.mgh.harvard.edu/)[20, 21], and the Desikan-Killiany Atlas[22] is used for dividing the brain into regions of interests: the procedures include skull stripping, registration, segmentation of white and gray matter, intensity normalization, tessellation of the gray-white matter boundary, automated topology correction, and surface deformation following intensity gradients to optimally place pial and white matter surface borders at the location where the greatest shift in intensity defines the transition to the other tissue class. The gray matter volume, which can serve as an indicative feature for distinguishing ASD group from NT group based on the previous research,[23, 24] is leveraged for classifying ASD. After processing the MRI data, we get 62-dimensional s-MRI volume-based features, indicating the gray matter volume at 62 different brain regions. However, according to some cross-section and longitudinal studies,[25-27] not all the brain regions are impacted by ASD. Thus, selecting ASD-related features can potentially improve the predictive accuracy accordingly.

**Genetic algorithm**. The genetic algorithm[28] is a population-based stochastic algorithm, which is employed here for feature selection. We first randomly initialize a population, whose size is controlled by the variable *N_pop* (**Table 1**). An individual in this population is a 62-dimensional binary vector (e.g., [1, 0, 0, 0, 1, …, 0, 1]), which plays the role of feature selector, i.e., we will select the features based on the indexes whose corresponding elements are '1s' in the individual. The selected features are then used to classify the ASD from the NT, and the predictive accuracy is used as the metric to evaluate the quality of the feature selector. Random forest[29] is used to quantify the relationship between the features and the binary labels (ASD/NT), and the predictive accuracy is calculated as the ratio of correctly classified labels and measured in 3-fold cross-validation. The number of trees is set to 100 in random forest. After initializing the population, we select individuals for crossover, which is based on the tournament selection.[30] The parent binary vectors are randomly chosen from the selected population for crossover (the crossover rate is controlled by *R_cross* (**Table 1**), e.g., if *R_cross* is 0.9, we cross any parent binary vectors under 90% of the time), which results in the offspring binary vectors. For example, if crossing [1, 1, 1, 1] and [0, 0, 0, 0], one example of offspring is [1, 1, 0, 0] and [0, 0, 1, 1]. We then randomly mutate parts of the elements in the offspring binary vectors, and the rate of mutation is controlled by *R_mut* (**Table 1**), e.g., if the *R_mut* is 1/62, then we mutate the elements in the offspring binary vector under 1/62 of the time, and the elements to be mutated are randomly selected. For example, if mutating one element of [1, 1, 0, 0], we can get [1, 1, 1, 0]. The goal of crossover and mutation is to obtain the best binary vector (i.e., feature selector) that lead to the highest predictive accuracy in classifying ASD. We can iterate this optimization process for several times, whose number is controlled by *N_iter* (**Table 1**). The random forest is implemented using Scikit-Learn,[31] and genetic algorithm is implemented from scratch using Python. The hyperparameters mentioned above can be tuned to make the best of the algorithm and we perform a hyperparameter study in **Table 1**, based on the UM1 data. According to this hyperparameter study, we could see that feature selection works for classifying ASD, as the predictive accuracy improves from 0.52 to 0.74. We finally choose the hyperparameter setting in Experiment 4 for latter study. Besides, based on the results in Experiment 4 and Experiment 5, longer optimizing time might not help find a feature combination that leads to higher predictive accuracy. Inspired by the hierarchical learning that long-horizon problems can be decomposed into simpler subproblems to solve,[32] we choose to do the feature selection hierarchically. Specifically, we do feature selection based on the features selected from the last-round-feature-selection (executing the feature selection based on the genetic algorithm and the hyperparameter setting chosen above is defined as one round of feature selection), until the predictive accuracy converges (APPENDIX A5). Noting that this is a heuristic method to get the optimal-feature-combination for classifying ASD, exploring the state-of-the-art method for feature selection is not the goal of this study. Since we apply such a heuristic method to data across all the sites consistently, the results from different sites are comparable to each other. We finally conduct three round of feature selection, the results of 1$^{st}$ to 3$^{rd}$ round feature selection are recorded in APPENDIX A2 to A4 respectively.

**Table 1.** Hyperparameter study in genetic algorithm based on UM1 data.

| **Experiment ID** | **N_iter** | **N_pop** | **R_cross** | **R_mut** | **Predictive accuracy** |
|---|---|---|---|---|---|
| 1 | 10 | 300 | 0.9 | 1/62 | 0.72±0.02 |
| 2 | 10 | 300 | 0.9 | 2/62 | 0.72±0.05 |
| 3 | 10 | 300 | 0.9 | 4/62 | 0.72±0.02 |
| 4 | 30 | 300 | 0.9 | 4/62 | 0.74±0.07 |
| 5 | 50 | 300 | 0.9 | 4/62 | 0.74±0.05 |
| original | - | - | - | - | 0.52±0.04 |

**Meta-data analysis**. After performing the hierarchical feature selection, the irrelevant features for classifying ASD are effectively eliminated. The predictive accuracies based on features selected in the last round are calculated for each site separately ($PRE_{last}$), whose results are recorded in APPENDIX A4. We then perform meta-data analysis, both quantitively and qualitatively based on Pearson correlation coefficient (r)[33] and t-SNE[34] visualization, to systematically investigate their impact on the predictive accuracy. Specifically, the $PRE_{last}$ are correlated to/visualized based on three kinds of meta-data (data size, phenotypic data of participants, scan parameters for collecting s-MRI) collected from each site, where detailed analysis of the correlation and visualized pattern are conducted. The illustration of the whole pipeline is shown in **Figure 1**.

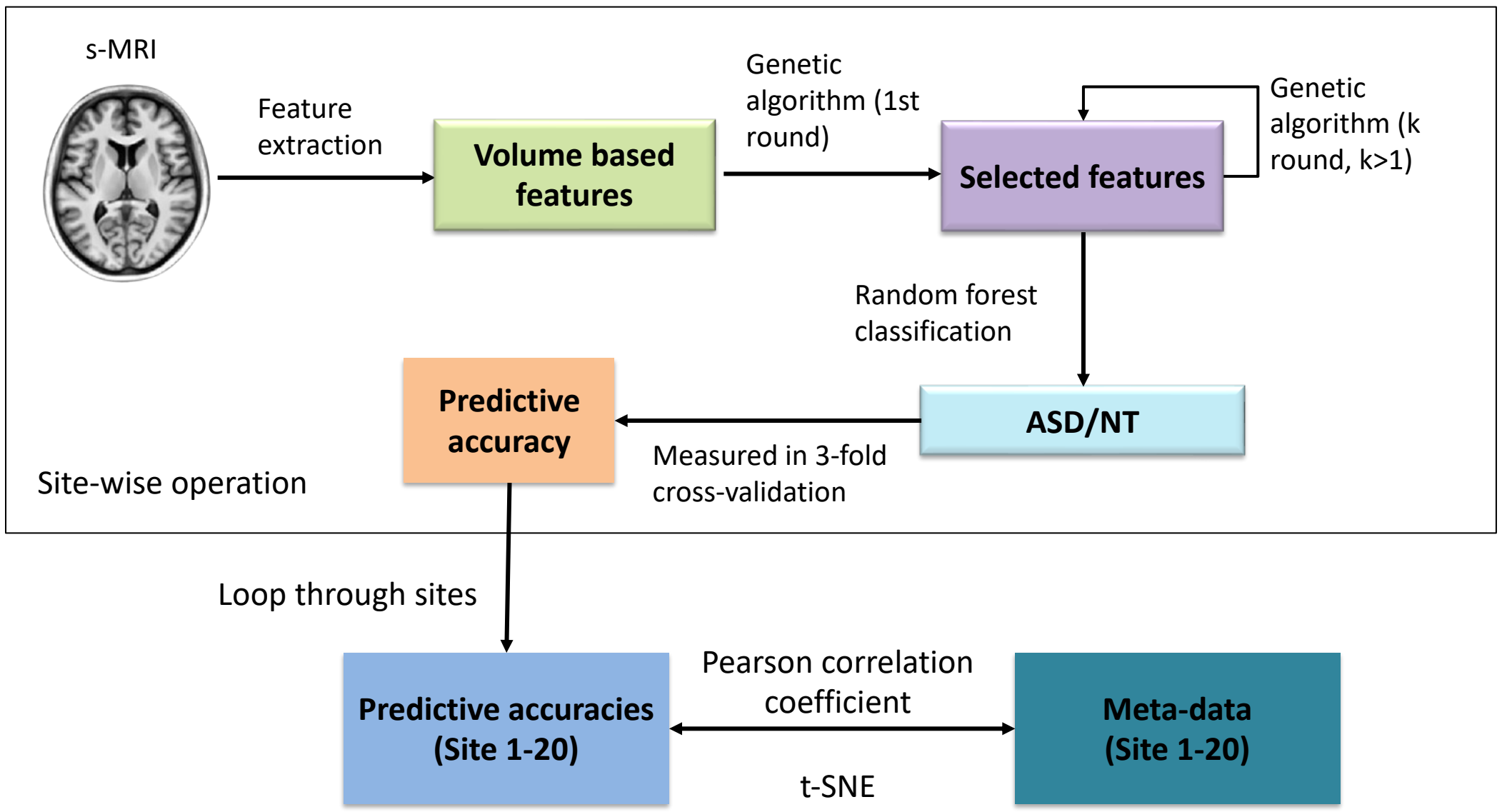

**Figure 1**. Pipeline of meta-data analysis.

## RESULTS

To explore meta-data's impact on the predictive accuracy of ASD classification, we calculate the site-wise predictive accuracy, since the data collected from different sites are heterogenous, due to different meta conditions for collecting the data. Before feature selection, the mean predictive accuracies in classifying ASD vary from 0.37 to 0.72 site by site (APPENDIX A1), indicating a heterogenous effect. After the hierarchical feature selection, the mean predictive accuracy over sites is improved from 0.53 to 0.85, indicating an effective elimination of the irrelevant features for classifying ASD. Besides, the variation in mean predictive accuracies (from 0.70 to 0.96, APPENDIX A4) persists even after the hierarchical feature selection, indicating such a heterogenous effect is invariant to feature selection. The visualization of predictive accuracies for 20 different sites, before and after hierarchical feature selection, is shown in **Figure 2**.

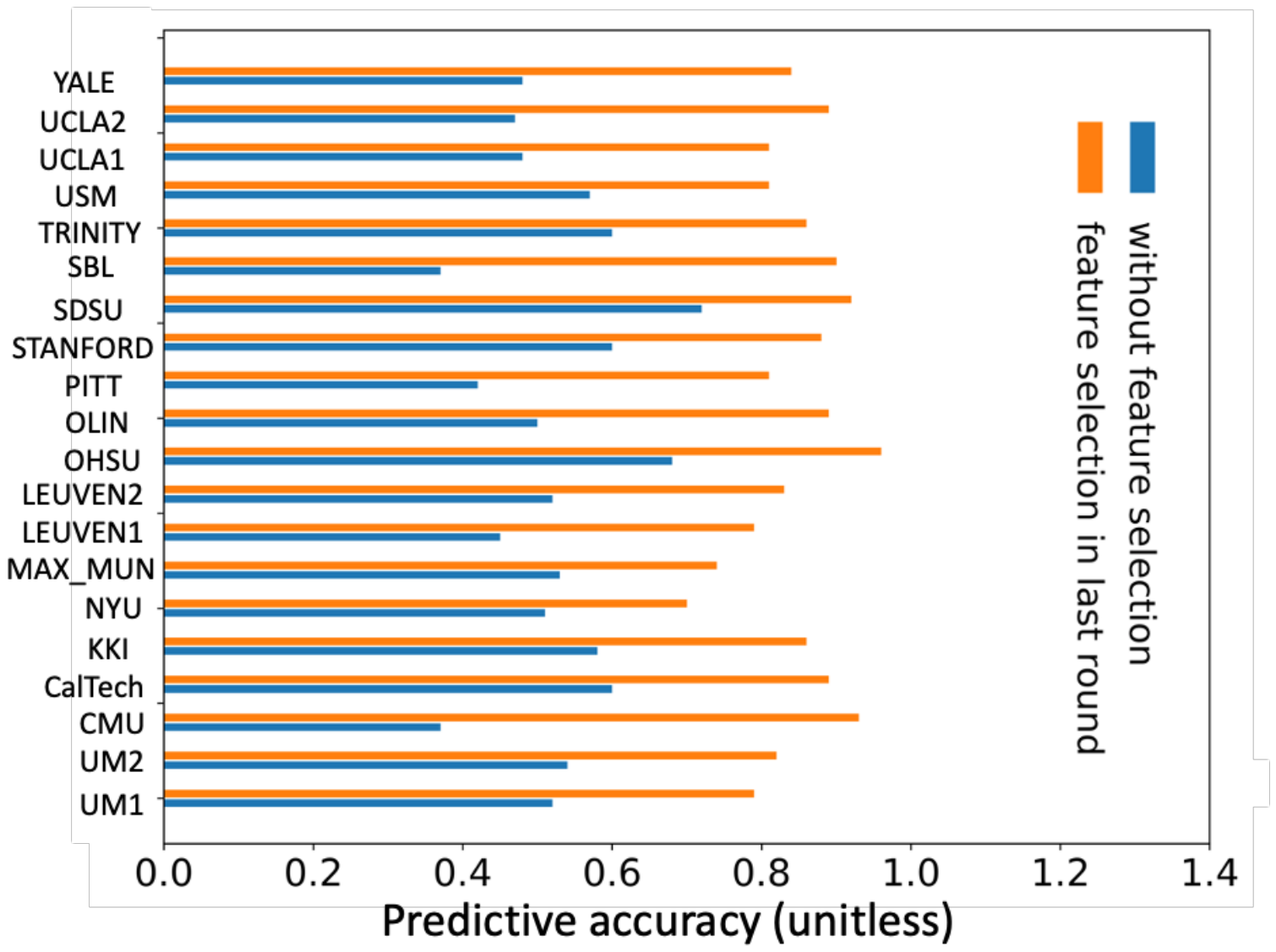

**Figure 2**. Predictive accuracies calculated for each site, under the scenario of without feature selection/feature selection in last round.

**Data size**. We first explore the impact of data size on the predictive accuracy for each site. As the feature selection goes on round by round, irrelevant features for classifying ASD are gradually eliminated, and the underlying relationship between s-MRI volume-based features and ASD should be more clearly uncovered. The Pearson correlation coefficient (r) is used to quantify the linear relationship between two variables. Based on the results in **Figure 3**, as irrelevant features are gradually eliminated (from a to d), a stronger and stronger negative relationship between the predictive accuracy and data size is observed, which is statistically significant (indicating by a p value << 0.05). Based on a survey[35] that performed statistical analysis of the data size's impacts on the classification accuracy of ASD based on neuroimaging data, the data size correlates negatively to the classification accuracy as well. However, they attributed the smaller dataset induced higher predictive accuracy to the inappropriate validation and testing methods used in machine learning. According to another work[36] that studied the MRI quality of ABIDE-I data from different sites, the data quality varied from site to site, although without indicating which site's quality was absolutely better than the other. Since we fix the machine learning algorithm and validation technologies on all the datasets involved, the data quality can be the only variation here. Noting that the relationship is quantified between ASD and s-MRI volume-based features, the variation of data quality is the direct factor causing the variation in the classification accuracy. Based on the results, simply adding more data is not a good strategy for improving the accuracy of classifying ASD, unless the data quality is strictly controlled.

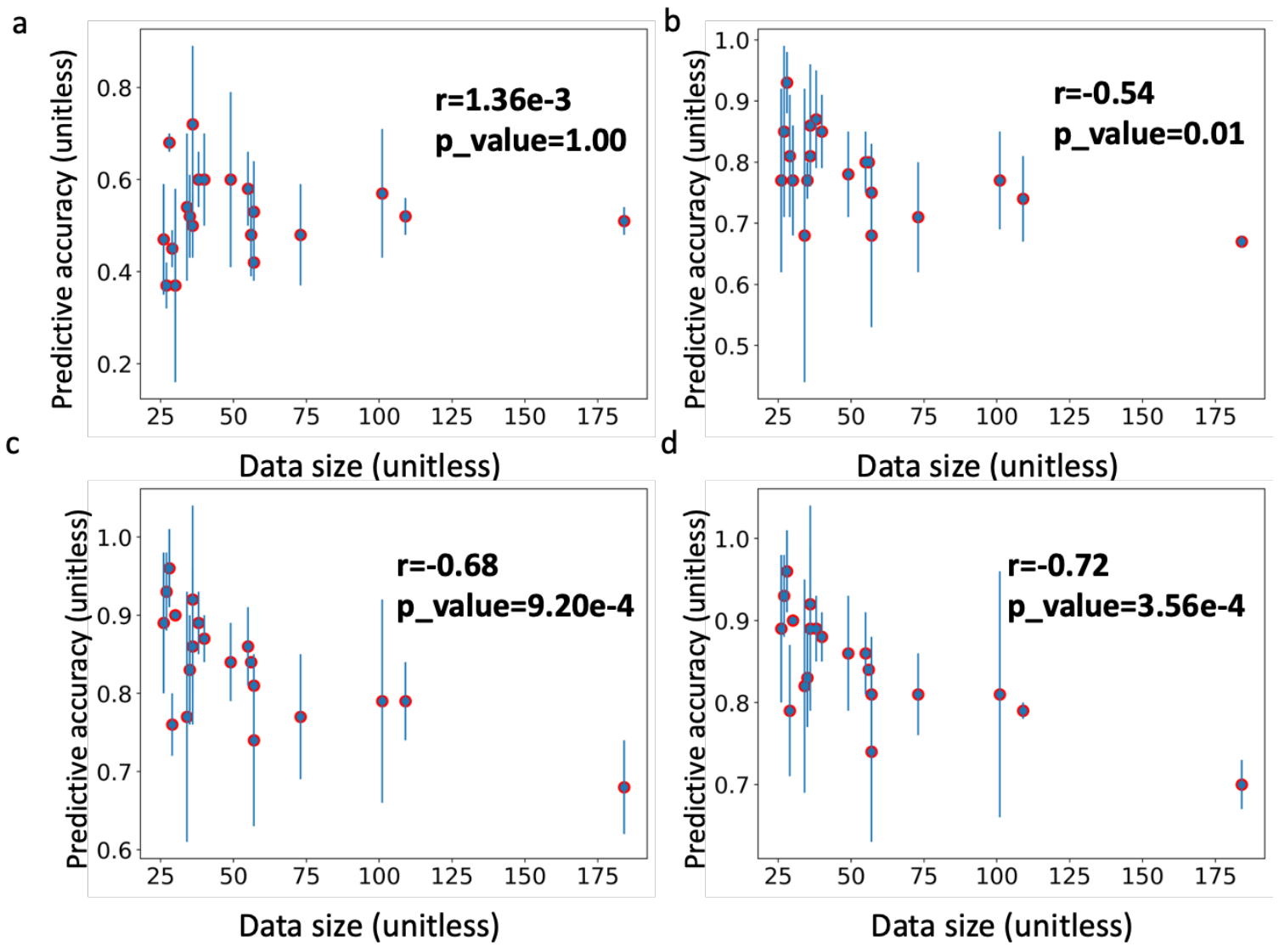

**Figure 3**. The pair plot between predictive accuracy of classifying ASD and the data size, a) without feature-selection, b) feature selection round 1, c) feature selection round 2, d) feature selection round 3. Each dot in each panel corresponds to one site.

**Phenotypic data of participants**. The s-MRI data are obtained from the participants involved in the study, who will influence the data quality directly. We analyze the phenotypic data provided in ABIDE-I, which inform what kinds of participants are involved in the study, and focus on the columns that have no missing and wrongly recorded values. The columns meeting the criteria are *AGE_AT_SCAN*, *SEX*, *EYE_STATUS_AT_SCAN*, which provide information about age at time of scan in years, subject gender, and eye status during rest of scan, respectively. We study the relationship between the statistics of those selected phenotypic data and site-wise predictive accuracy ($PRE_{last}$) in **Figure 4**. The mean and standard deviation of age (**Figure 4a-b**), the ratio of female versus male ($N_{female}/N_{male}$) (**Figure 4c**), and the median of the eye status (the eye status during the rest of scan is a categorical variable, with 1 indicating eyes open and 2 indicating eyes closed) (**Figure 4d**) are explored for the ASD group. Based on the results in APPENDIX A6, if simply correlating the predictive accuracy ($PRE_{last}$) to the statistics of phenotypic data of ASD participants within each site, the robust conclusion can't be arrived, indicating by high p-values ($\gg 0.05$). Thus, we conduct bootstrapping within each site to collect more samples for statistics, to potentially reduce the p-values. Specifically, 50% of the data within each site are randomly sampled for 50 times (based on the features selected in the last round), and the statistics of the phenotypic data and the predictive accuracy of classifying ASD based on the sampled data are calculated then. The pair plot of the bootstrapped predictive accuracy and bootstrapped statistics of the phenotypic data are shown in **Figure 4**, and the corresponding r and p-value are calculated. Based on the results, a weak correlation is shown in all the plots, indicating that age at time of scan, subject gender, and eye status during rest of scan have insignificant impact on the predictive accuracy of ASD classification, indicating that the different phenotype has insignificant influence on the structure of brain regions that are affected with ASD. Previous research on the impact of age, sex and eye status in ASD detection have also been investigated. For example, in ref. [19], the ASD classification was conducted in adult group and adolescent group separately based on neuroimaging data, where only males were involved, and the predictive accuracies were different for different groups. Based on their results, the brain regions that were affected with ASD were different for different age-groups, indicating that the features required for classifying ASD were different for different age-groups. However, according to our results, such an influence might be trivial, as indicated by a small r (< 0.3). In ref. [37], the sex and gender difference in ASD has been investigated. Based on their discussion, the accuracies of ASD detection in different gender might vary, which can be attributed to the different phenotype and genetic information in different gender. However, according to our research, gender plays a non-critical role in ASD classification that is based on brain structural information. In ref. [38], eye gaze was used as a biomarker for ASD detection, where the eye status plays a critical role in detecting ASD. However, according to our results, the eye status won't make any effective change to the brain regions that are affected with ASD.

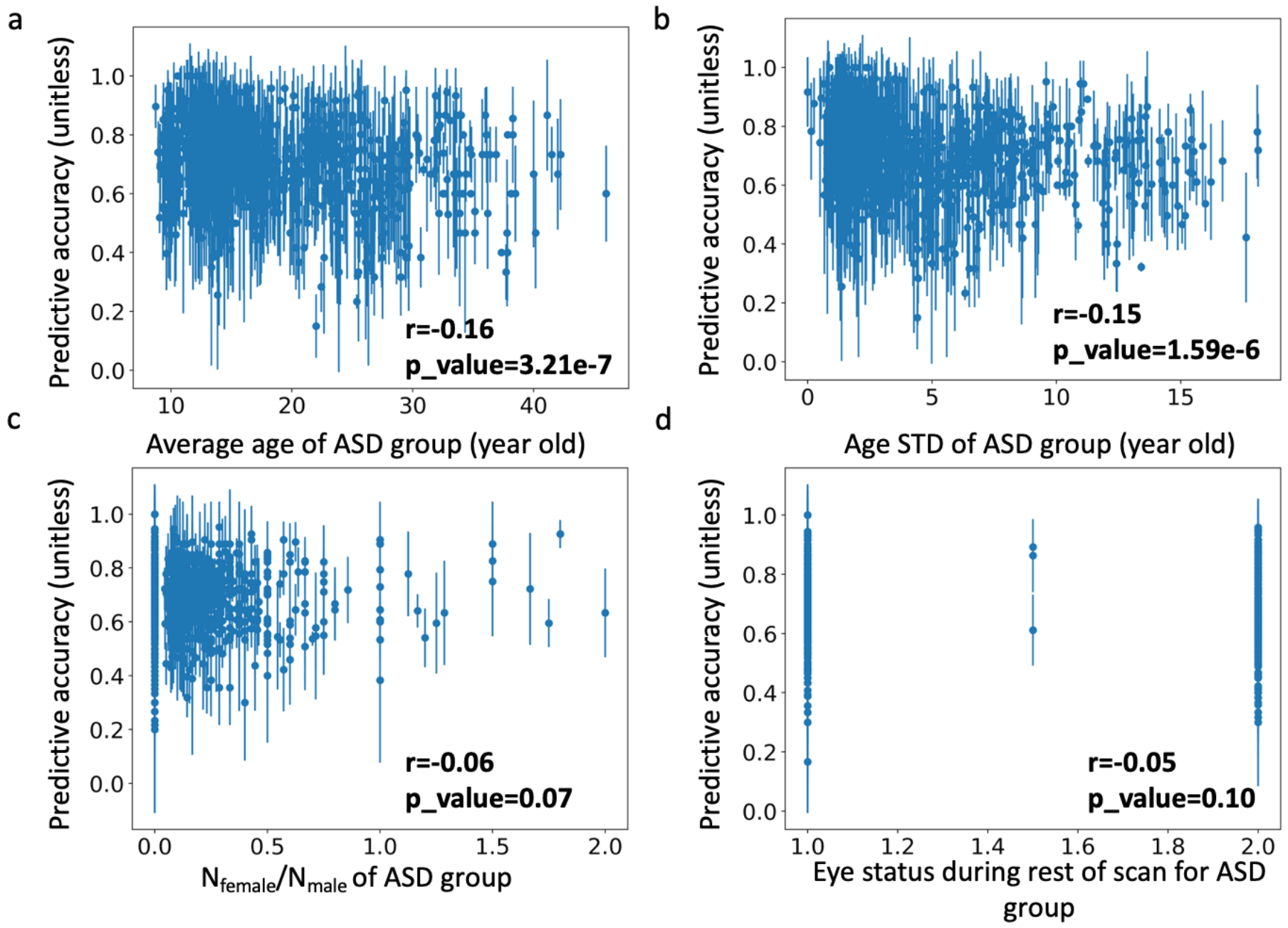

**Figure 4**. The pair plot between predictive accuracy of classifying ASD and the statistics of phenotypic data of the ASD participants within each site, a) average age of ASD group, b) age STD of ASD group, c) ratio of female versus male of ASD group, d) eye status during rest of scan for ASD group.

**Scan parameters for collecting s-MRI**. Finally, we explore the scan parameters' impacts on the predictive accuracy. The full details of scan parameters are recorded on the website of ABIDE-I, in the format of both text and numerical values. Exploring all the scan parameters is insurmountable as there is a great amount of information in them. Thus, we focus on investigating the information that might have non-trivial impact on the quality of MRI based on ref. [36], and we record those scan parameters in **Table 2**. *MRI Scanner Vendor* indicates where the scanner is manufactured; echo times (TE) and repetition times (TR) represent the time of recover and decay in magnetization before measuring the MR single; the flip-angle (FA) is associated with the contrast weighting of MRI; and the total scan duration (TI) reflects the MRI resolution. Since the scan parameters are integrated for practical use, exploring the impact of single parameter on predictive accuracy is meaningless. For *TR* and *TI* columns, there are null values in them, as indicated by *NA*, we will abandon these two columns in latter analysis. The categorical variables in *MRI Scanner Vendor* columns are first encoded into a numerical vector [1, 1, 6, 5, 2, 4, 6, 3, 3, 5, 4, 4, 1, 0, 3, 2, 4, 5, 5], with each integer representing a category. Then it is concatenated with *TE* and *FA*, as a representation of the scan conditions used for collecting MRI. In **Figure 5**, the 3-d vectors from all the sites are then projected onto the 2-d space using t-SNE[34] and labeled by the corresponding predictive accuracies ($PRE_{last}$). Based on the results, the predictive accuracies can vary significantly (from ~0.70 to ~0.95) even the scan conditions used for collecting s-MRI are similar, as indicated by red circle in the top left corner. However, very different scan conditions (indicated by a large distance between the two black circles) can lead to similar predictive accuracies. So, the scan parameters solely don't have a clear implication on whether it will produce the s-MRI in high quality, i.e., if the scan parameters used in one site help produce s-MRI in high quality, it does not mean other sites can produce s-MRI at equivalent quality by using the same scan parameters.

**Table 2**. Records of scan parameters that can influence the MRI quality.

| **Site** | **MRI Scanner Vendor** | **TR (sec)** | **TE (sec)** | **TI (sec)** | **FA (deg)** |
|---|---|---|---|---|---|
| Caltech | Siemens Magnetom TrioTim | 1.59 | 2.73e-3 | 0.80 | 10 |
| CMU | Siemens Magnetom Verio | 1.87 | 2.48e-3 | 1.10 | 8 |
| KKI | Philips Achieva 3T | 8.00e-3 | 3.70e-3 | 0.80 | 8 |
| MAXMUN | Siemens Magnetom Verio | 1.80 | 3.06e-3 | 0.90 | 9 |
| NYU | Siemens Magnetom Allegra | 2.53 | 3.25e-3 | 1.10 | 7 |
| OLIN | Siemens Magnetom Allegra | 2.50 | 2.74e-3 | 0.90 | 8 |
| OHSU | Siemens Magnetom TrioTim | 2.30 | 3.58e-3 | 0.90 | 10 |
| SDSU | General Electric Discovery MR750 3T | 11.10e-3 | 4.30e-3 | 0.60 | 8 |
| SBL | Philips Intera 3T | 9.00e-3 | 3.50e-3 | NA | 7 |
| STANFORD | General Electric Signa 3T | 8.40e-3 | 1.80e-3 | NA | 15 |
| TRINITY | Philips Achieva 3T | 8.50e-3 | 3.90e-3 | 1.00 | 8 |
| UCLA1 | Siemens Magnetom TrioTim | 2.30 | 2.84e-3 | 0.85 | 9 |
| UCLA2 | Siemens Magnetom TrioTim | 2.30 | 2.84e-3 | 0.85 | 9 |
| LEUVEN1 | Philips Intera 3T | 9.60e-3 | 4.60e-3 | 0.90 | 8 |
| LEUVEN2 | Philips Intera 3T | 9.60e-3 | 4.60e-3 | 0.90 | 8 |
| UM1 | General Electric Signa 3T | NA | 1.80e-3 | NA | 15 |
| UM2 | General Electric Signa 3T | NA | 1.80e-3 | NA | 15 |
| PITT | Siemens Magnetom Allegra | 2.10 | 3.93e-3 | 1.00 | 7 |
| USM | Siemens Magnetom Allegra | 2.10 | 3.93e-3 | 1.00 | 7 |
| YALE | Siemens Magnetom TrioTim | 1.23 | 1.73e-3 | 0.60 | 9 |

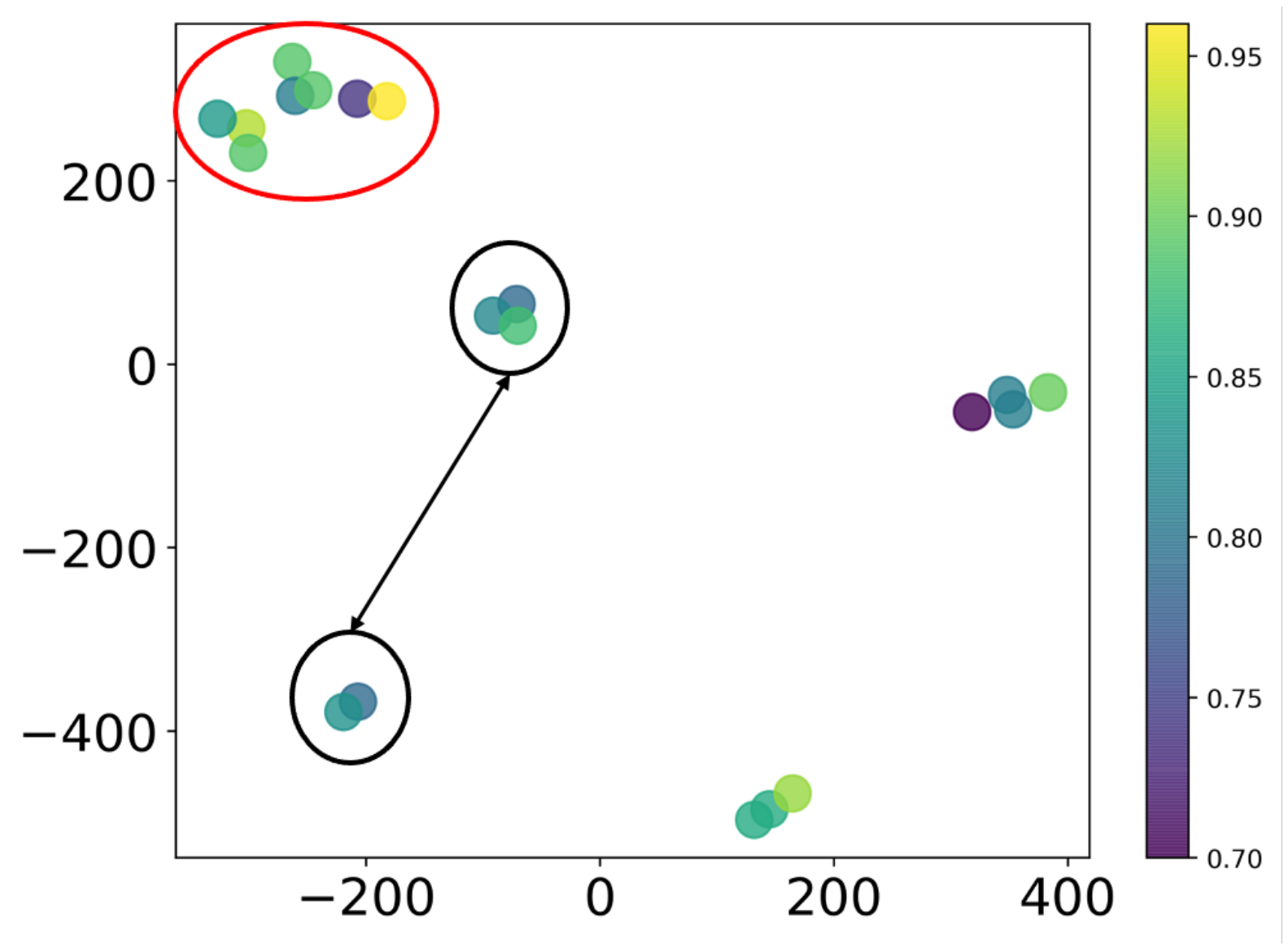

**Figure 5**. The visualization of the representation of scan conditions from each site, labeled by the predictive accuracy ($PRE_{last}$) from each site.

## CONCLUSION

In summary, we systematically investigate three kinds of meta-data's impact on the predictive accuracy of ASD classification. Hierarchical feature selection based on genetic algorithm is conducted before the meta-data analysis, to effectively eliminate irrelevant features for ASD classification. A negative relationship is observed between predictive accuracy and data size, indicating that simply increasing the number of data won't help improve the predictive accuracy, unless the data quality can be strictly controlled. Phenotypic data of ASD participants, like age, gender, and eye status during the rest of scan, are shown to have insignificant impact on the predictive accuracy, indicating that phenotype have insignificant influence on the structure of brain region that are affected with ASD. The scan parameters' impact on predictive accuracy is finally investigated, showing that scan parameters can't be considered as a factor to improve the quality of MRI, and thus the predictive accuracy, solely.

## ACKNOWLEDGEMENTS

The authors acknowledge the financial support from Shenzhen Science and Technology Program (No. KQTD20200820113106007), and National Science Foundation of China (Grant No. 81171488, 81671669 and 81820108018). The computation was supported by the Sugon computing platform under the grant KQTD20200820113106007.

## CODE AVAILABILITY

Code and data will be available at https://github.com/RUIMINMA1996/ASD_GM upon the publishment of this work.

## APPENDIX

**A1**. Predictive accuracy of classifying ASD based on all 62 features, based on data from different sites.

| Site name | Data size | Predictive accuracy | Site name | Data size | Predictive accuracy |
|---|---|---|---|---|---|
| UM1 | 109 | 0.52±0.04 | OLIN | 36 | 0.50±0.07 |
| UM2 | 34 | 0.54±0.16 | PITT | 57 | 0.42±0.04 |
| CMU | 27 | 0.37±0.05 | STANFORD | 40 | 0.60±0.10 |
| CalTech | 38 | 0.60±0.06 | SDSU | 36 | 0.72±0.17 |
| KKI | 55 | 0.58±0.08 | SBL | 30 | 0.37±0.21 |
| NYU | 184 | 0.51±0.03 | TRINITY | 49 | 0.60±0.19 |
| MAX_MUN | 57 | 0.53±0.11 | USM | 101 | 0.57±0.14 |
| LEUVEN1 | 29 | 0.45±0.04 | UCLA1 | 73 | 0.48±0.11 |
| LEUVEN2 | 35 | 0.52±0.09 | UCLA2 | 26 | 0.47±0.12 |
| OHSU | 28 | 0.68±0.02 | YALE | 56 | 0.48±0.09 |

**A2**. Predictive accuracy of classifying ASD based on the features selected by genetic algorithm in round 1, using data collected from different sites separately.

| **Site name** | **Data size** | **Predictive accuracy** | **Site name** | **Data size** | **Predictive accuracy** |
|---|---|---|---|---|---|
| UM1 | 109 | 0.74±0.07 | OLIN | 36 | 0.81±0.04 |
| UM2 | 34 | 0.68±0.24 | PITT | 57 | 0.75±0.07 |
| CMU | 27 | 0.85±0.14 | STANFORD | 40 | 0.85±0.06 |
| CalTech | 38 | 0.87±0.08 | SDSU | 36 | 0.86±0.10 |
| KKI | 55 | 0.80±0.05 | SBL | 30 | 0.77±0.09 |
| NYU | 184 | 0.67±0.01 | TRINITY | 49 | 0.78±0.07 |
| MAX_MUN | 57 | 0.68±0.15 | USM | 101 | 0.77±0.08 |
| LEUVEN1 | 29 | 0.81±0.10 | UCLA1 | 73 | 0.71±0.09 |
| LEUVEN2 | 35 | 0.77±0.03 | UCLA2 | 26 | 0.77±0.15 |
| OHSU | 28 | 0.93±0.05 | YALE | 56 | 0.80±0.05 |

**A3**. Predictive accuracy of classifying ASD based on the features selected by genetic algorithm in round 2, using data collected from different sites separately.

| **Site name** | **Data size** | **Predictive accuracy** | **Site name** | **Data size** | **Predictive accuracy** |
|---|---|---|---|---|---|
| UM1 | 109 | 0.79±0.05 | OLIN | 36 | 0.86±0.10 |
| UM2 | 34 | 0.77±0.16 | PITT | 57 | 0.81±0.02 |
| CMU | 27 | 0.93±0.05 | STANFORD | 40 | 0.87±0.03 |
| CalTech | 38 | 0.89±0.04 | SDSU | 36 | 0.92±0.12 |
| KKI | 55 | 0.86±0.05 | SBL | 30 | 0.90±0.00 |
| NYU | 184 | 0.68±0.06 | TRINITY | 49 | 0.84±0.05 |
| MAX_MUN | 57 | 0.74±0.11 | USM | 101 | 0.79±0.13 |
| LEUVEN1 | 29 | 0.76±0.04 | UCLA1 | 73 | 0.77±0.08 |
| LEUVEN2 | 35 | 0.83±0.07 | UCLA2 | 26 | 0.89±0.09 |
| OHSU | 28 | 0.96±0.05 | YALE | 56 | 0.84±0.00 |

**A4**. Predictive accuracy of classifying ASD based on the features selected by genetic algorithm in round 3, using data collected from different sites separately.

| Site name | Data size | Predictive accuracy | Site name | Data size | Predictive accuracy |
|---|---|---|---|---|---|
| UM1 | 109 | 0.79±0.01 | OLIN | 36 | 0.89±0.10 |
| UM2 | 34 | 0.82±0.13 | PITT | 57 | 0.81±0.07 |
| CMU | 27 | 0.93±0.05 | STANFORD | 40 | 0.88±0.03 |
| CalTech | 38 | 0.89±0.04 | SDSU | 36 | 0.92±0.12 |
| KKI | 55 | 0.86±0.05 | SBL | 30 | 0.90±0.00 |
| NYU | 184 | 0.70±0.03 | TRINITY | 49 | 0.86±0.07 |
| MAX_MUN | 57 | 0.74±0.11 | USM | 101 | 0.81±0.15 |
| LEUVEN1 | 29 | 0.79±0.08 | UCLA1 | 73 | 0.81±0.05 |
| LEUVEN2 | 35 | 0.83±0.06 | UCLA2 | 26 | 0.89±0.09 |
| OHSU | 28 | 0.96±0.05 | YALE | 56 | 0.84±0.00 |

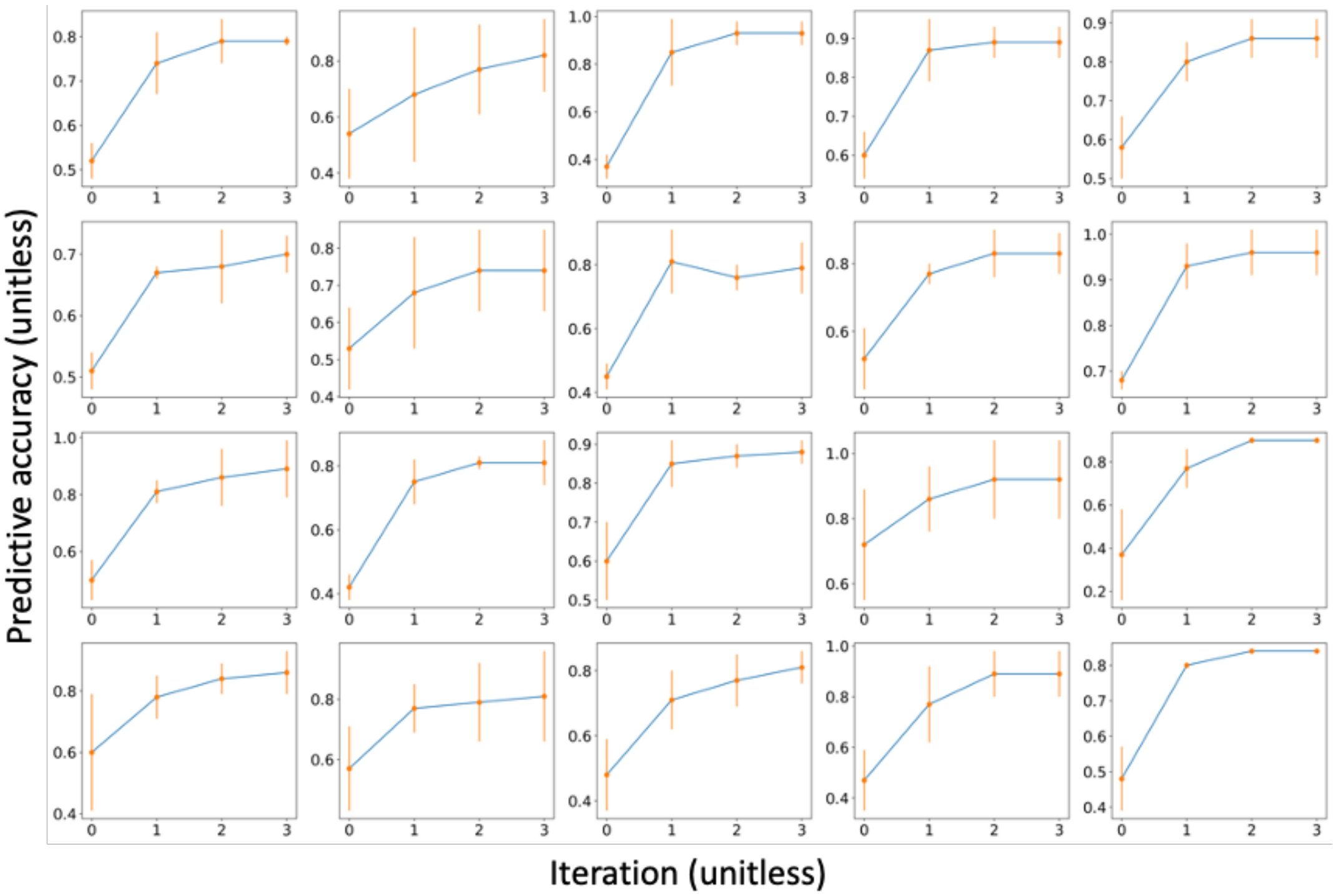

**A5**. The predictive accuracy of classifying ASD in each round of feature selection, each panel corresponds to a site.

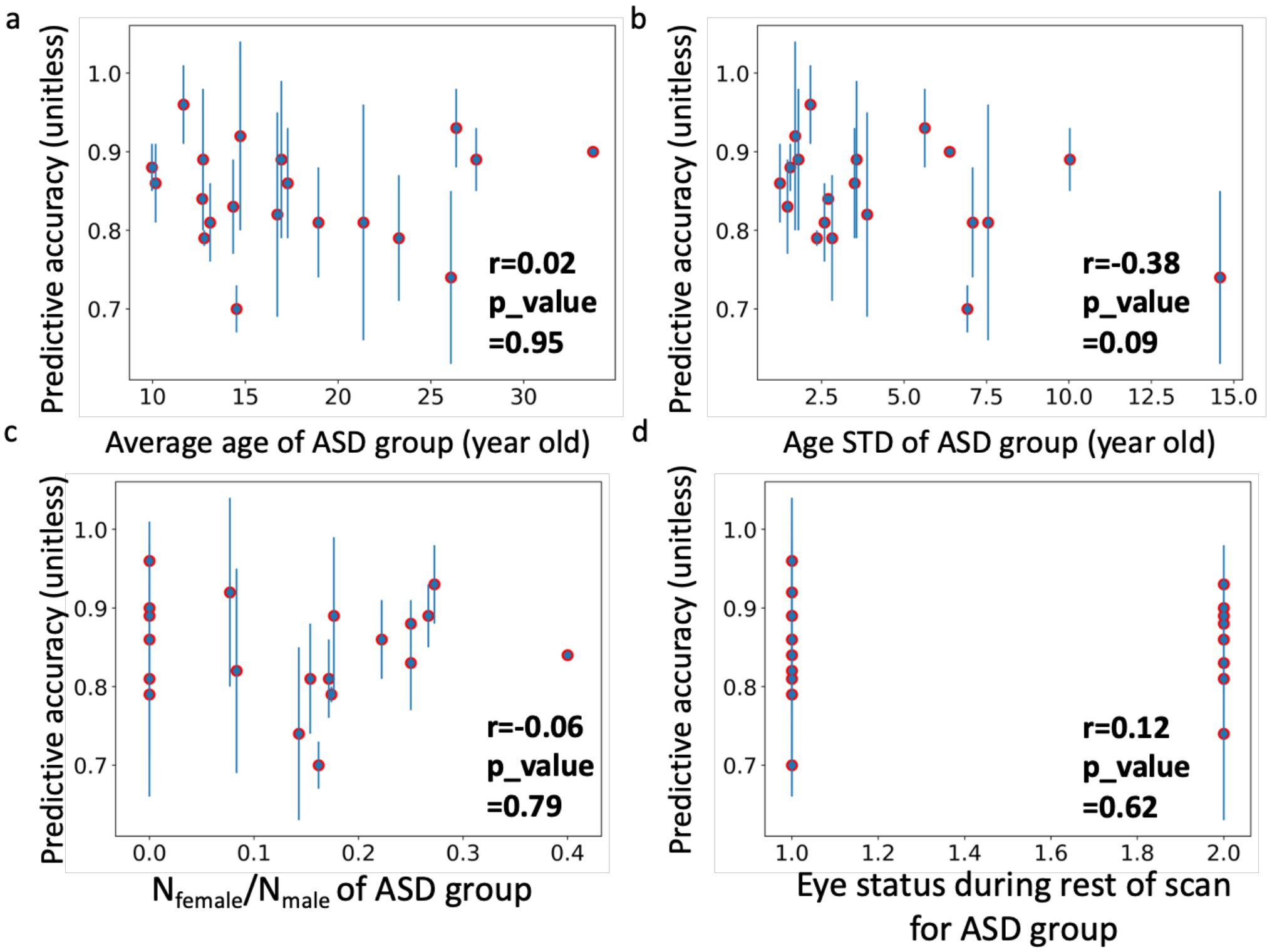

**A6**. The pair plot between predictive accuracy of classifying ASD and the statistics of phenotypic data of the ASD participants within each site, a) average age of ASD group, b) age STD of ASD group, c) ratio of female versus male of ASD group, d) eye status during rest of scan for ASD group.